\documentclass[runningheads]{llncs}
\usepackage{graphicx}
\usepackage{amsmath,amssymb} 
\usepackage{tabularx}
\usepackage{color}
\usepackage{url}
\usepackage[width=122mm,left=12mm,paperwidth=146mm,height=193mm,top=12mm,paperheight=217mm]{geometry}

\newcommand{\bs}[1]{\ensuremath{\boldsymbol{#1}}}
\newcommand{\tsup}{\textsuperscript}
\DeclareMathOperator* {\argmax}{arg\, max}

\begin{document}
\pagestyle{headings}
\mainmatter
\def\ECCV16SubNumber{}
\title{A Focused Dynamic Attention Model for Visual Question Answering}
\titlerunning{A Focused Dynamic Attention Model for Visual Question Answering}

\authorrunning{Ilija Ilievski, Shuicheng Yan, Jiashi Feng}

\author{Ilija Ilievski, Shuicheng Yan, Jiashi Feng\\
\texttt{\small ilija.ilievski@u.nus.edu, \{eleyans,elefjia\}@nus.edu.sg}}
\institute{National University of Singapore}

\maketitle

\begin{abstract}
Visual Question and Answering (VQA) problems are attracting increasing interest from multiple research disciplines.
Solving VQA problems requires techniques from both computer vision for understanding the visual contents of a presented image or video, as well as the ones from natural language processing for understanding semantics of the question and generating the answers.
Regarding visual content modeling, most of existing VQA methods adopt the strategy of extracting global features from the image or video, which inevitably fails in capturing fine-grained information such as spatial configuration of multiple objects.
Extracting features from auto-generated regions -- as some region-based image recognition methods do -- cannot essentially address this problem and may introduce some overwhelming irrelevant features  with the question.
In this work, we propose a novel Focused Dynamic Attention (FDA) model to provide better aligned image content representation with proposed questions.
Being aware of the key words in the question, FDA employs off-the-shelf object detector to identify important regions and fuse the information from the regions and global features via an LSTM unit.
Such question-driven representations are then combined with question representation and fed into a reasoning unit for generating the answers.
Extensive evaluation on a large-scale benchmark dataset, VQA, clearly demonstrate the superior performance of FDA over well-established baselines.
   \keywords{Visual Question Answering, Attention}
\end{abstract}

\section{Introduction} 

Visual question answering (VQA) is an active research direction that lies in the intersection of computer vision, natural language processing, and machine learning. 
Even though with a very short history, it already has received great research attention from multiple communities. 
Generally, the VQA investigates a generalization  of traditional  QA problems where visual input (\emph{e.g.}, an image) is necessary to be  considered.
More concretely, VQA is about  how to provide a correct answer to a human posed question concerning contents of one presented image or video.

VQA is a quite challenging task and undoubtedly important for developing modern AI systems.
The VQA problem can be regarded as a Visual Turing Test \cite{geman-15,malinowski-14}, and besides contributing to the advancement of the involved research areas, it has other important applications, such as blind person assistance and image retrieval.
Coming up with solutions to this task requires natural language processing techniques for understanding the questions and generating the answers,  as well as computer vision techniques for understanding contents of the concerned image.
With help of these two core techniques, the computer can perform reasoning about the perceived contents and  posed questions.

Recently, VQA is advanced significantly by the development of machine learning methods (in particular the deep learning ones) that can learn proper representations of questions and images, align and fuse them in a joint question-image space and  provide a direct mapping from this joint representation to a correct answer.

For example, consider the following image-question pair:  an image of an apple tree with a basket of apples next to it, and a question ``How many apples are in the basket?''.
Answering this question requires VQA methods to first understand  the semantics of the question, then locate the objects (apples) in the image, understand the relation between the image objects (which apples are in the basket), and finally count them and generate an answer with the correct number of apples.

The first feasible solution to VQA problems  was provided by Malinowski and Fritz in \cite{malinowski-14}, where they used a semantic language parser and a Bayesian reasoning model, to understand the meaning of questions and to generate the proper answers.
Malinowski and Fritz also constructed the first VQA benchmark dataset, named as DAQUAR, which contains 1,449 images and 12,468 questions generated by humans or automatically by following a template and extracting facts from a database \cite{malinowski-14}.
Shortly after, Ren et al. \cite{ren-15} released the TORONTO--QA dataset, which contains a large number of images (123,287) and questions (117,684), but the questions are automatically generated and thus can be answered without complex reasoning.
Nevertheless, the release of the TORONTO--QA dataset was important since it provided enough data for deep learning models to be trained and evaluated on the VQA problem \cite{gao-15,malinowski-15,ren-15}.
More recently, Antol et al.~\cite{antol-15} published the currently largest VQA dataset.
It consists of three human posed questions and ten answers given by different human subjects, for each one of the 204,721 images found in the Microsoft COCO dataset \cite{lin-14}.
Answering the 614,163 questions requires complex reasoning, common sense, and real-world knowledge, making the VQA dataset suitable for a true Visual Turing Test.
The VQA authors split the evaluation on their dataset on two tasks: an open-ended task, where the method should generate a natural language answer, and a multiple-choice task, where for each question the method should chose one of the 18 different answers. 

The current top performing methods \cite{noh-15,wu-15,ma-15} employ deep neural network model that predominantly uses the convolutional neural network (CNN) architecture \cite{lecun-89,krizhevsky-12,simonyan-14,szegedy-15} to extract image features and a Long Short-Term Memory (LSTM) \cite{hochreiter-97} network to extract the representations for  questions.
The CNN and LSTM representation vectors are then usually fused by concatenation \cite{zhou-15,ren-15,malinowski-15} or element-wise multiplication \cite{shih-15,chen-15}.
Other approaches additionally incorporate some kind of attention mechanism over the image features \cite{chen-15,yang-15,xu-15}.

Properly modeling the image contents is one of the critical factors for solving VQA problems well.
A common practice with  existing VQA methods on modeling image contents is to extract  global features for the overall image.
However,  only using global feature is arguably insufficient to capture all the necessary visual information and provide full understanding of image contents such as multiple objects, spatial configuration of the objects and informative background. 
This issue can be relieved to some extent by extracting features from object proposals -- the image regions that possibly contain objects of interest. 
However, using  features from all image regions \cite{yang-15,chen-15}  may provide too much noise or overwhelming  information irrelevant to the question and thus hurt the overall VQA performance.

In this work, we propose a question driven attention  model that is able to automatically identify and focus on image regions  relevant for the current question.
We name our proposed model Focused Dynamic Attention (FDA) for Visual Question Answering.
With the FDA model, computers can select and recognize the image regions  in a well-aligned sequence with the key words containing in a given question.
Recall the above VQA example.
To answer the question of ``How many apples are in the basket?'', FDA would first localize the regions corresponding to the key words ``apples'' and ``basket'' (with the help of a generic object detector) and  extract description features from these regions of interest.
Then VQA  compliments the features from selected image regions with a global image feature providing contextual information for the overall image, and reconstruct a visual representation by encoding them with a Long Short-Term Memory (LSTM) unit. 
 
We evaluate and compare the performance of our proposed FDA model on two types of VQA tasks, \emph{i.e.}, the open-ended task and the multiple-choice task, on the  VQA dataset -- the largest VQA benchmark dataset.
Extensive experiments demonstrate that FDA brings substantial performance improvement upon well-established baselines.

The main contributions of this work can be summarized as follows:
\begin{itemize}
  \item We introduce a focused dynamic attention mechanism that learns to use the question word order to shift the focus from one image object, to another. 
  \item We describe a model that fuses local and global context visual features with textual features.
  \item We perform an extensive evaluation, comparing to all existing methods, and achieve state-of-the-art accuracy on the open-ended, and on the multiple-choice VQA tasks.
\end{itemize}

The rest of the paper is organized as follows.
In Section~\ref{sec:related} we review the current VQA models, and compare them to our model. 
We formulate the problem and explain our motivation in Section~\ref{sec:method_overview}. 
We describe our model in Section~\ref{sec:model} and in Section~\ref{sec:experiments} we evaluate and compare it with the current state-of-the-art models.
We conclude our work in Section~\ref{sec:conclusion}.

\section{Related Work}\label{sec:related}
VQA has received great research attention recently and a couple of methods have been developed to solve this problem.
The  most similar model to ours is  the Stacked Attention Networks (SAN) proposed by Yang et al. \cite{yang-15}. 
Both models use attention mechanism that combines the words and image regions.
However, \cite{yang-15} use convolutional neural network to put attention over the image regions, based on the question word unigrams, bigrams, and trigrams.
Further, their attention mechanism is not using object bounding boxes, which makes the attention less focused. 

Another model that uses attention mechanism in solving VQA problems is the ABC-CNN model described in \cite{chen-15}.
ABC-CNN uses the question embedding to configure convolutional kernels that will define an attention weighted map over the image features.
The advantage of our FDA model over ABC-CNN is two fold. 
First, FDA employs an LSTM network to encode the image region features in a order that corresponds to the question word order.
Second, FDA does not put handcrafted weights on the image features (a practice showed to hurt the learning process in our experiments). Instead, FDA extracts CNN features directly from the cropped image regions of interest. In this sense, FDA is  more efficient than ABC-CNN in visual contents modeling.

Yet another attention model for visual question answering is proposed in \cite{shih-15}. 
The work, is closely related to the work by \cite{chen-15}, in that it also applies a weighted map over the image and the question word features.  
However, similar to our work, they use object proposals from \cite{zitnick-14} to select image regions instead of the whole image. 
Different from that work, our proposed FDA model also employs the information embedded in the order of the question words and  focuses  on the corresponding object bounding boxes.
In contrast, the model proposed in \cite{zitnick-14} straightforwardly concatenate all the image region features with the question word features and feed them all at once to a two layer network. 

Jiang et al. propose another model that combines the CNN image features and an LSTM network for encoding the multimodal representation, with the addition of a Compositional Memory units which fuse the image and word feature vectors \cite{jiang-15}.

Ma et al. in \cite{ma-15} take an interesting approach and use three convolutional neural networks to represent not only the image, but also the question, and their common representation in a multimodal space.
The multimodal representation is then fed to a SoftMax layer to produce the answer.  

Another interesting approach worth mentioning is the work by Andreas et al.~\cite{andreas-16}.
They use a semantic grammar parser to parse the question and propose neural network layouts accordingly.
They train a model to learn to compose a network from one of the proposed network layouts using several types of neural modules, each specifically designed to address the different sub-tasks of the VQA problem (e.g. counting, locating an object, etc.).

\section{Method Overview}\label{sec:method_overview}
In this section, we briefly describe the motivation and give formal problem formulation.

\subsection{Problem Formulation}

The visual question answering problem can be represented as predicting the best answer $\hat{a}$ given an image $I$ and a question $q$.
Common practice~\cite{antol-15,zhou-15,ren-15,yang-15} is to use the 1,000 most common answers in the training set and thus simplify the VQA task to a classification problem.
The following equation represents the problem mathematically:

\begin{equation}
   \bs{\hat{a}}=\argmax_{\bs{a}\in \Omega}p(\bs{a}|\bs{I},\bs{q};\bs{\theta})
\end{equation}
where $\Omega$ is the set of all possible answers and $\theta$ are the model weights. 

\subsection{Motivation} 
The baseline methods from \cite{antol-15} show only modest increase in accuracy when including the image features (4.98\% for open-ended questions, and 2.42\% for multiple-choice question). 
We believe that the image contains a lot more information and should increase the accuracy much more. 
Thus, we focus on improving the image features and design a visual attention mechanism, which learns to focus on the question related image regions. 

The proposed attention mechanism is loosely inspired on the human visual attention mechanism. 
Humans shift the focus from one image region to another, before understanding how the regions relate to each other and grasping the meaning of the whole image.
Similarly, we feed our model image regions relevant for the question at hand, before showing the whole image.

\begin{figure}[!ht] 
\includegraphics[width=\linewidth]{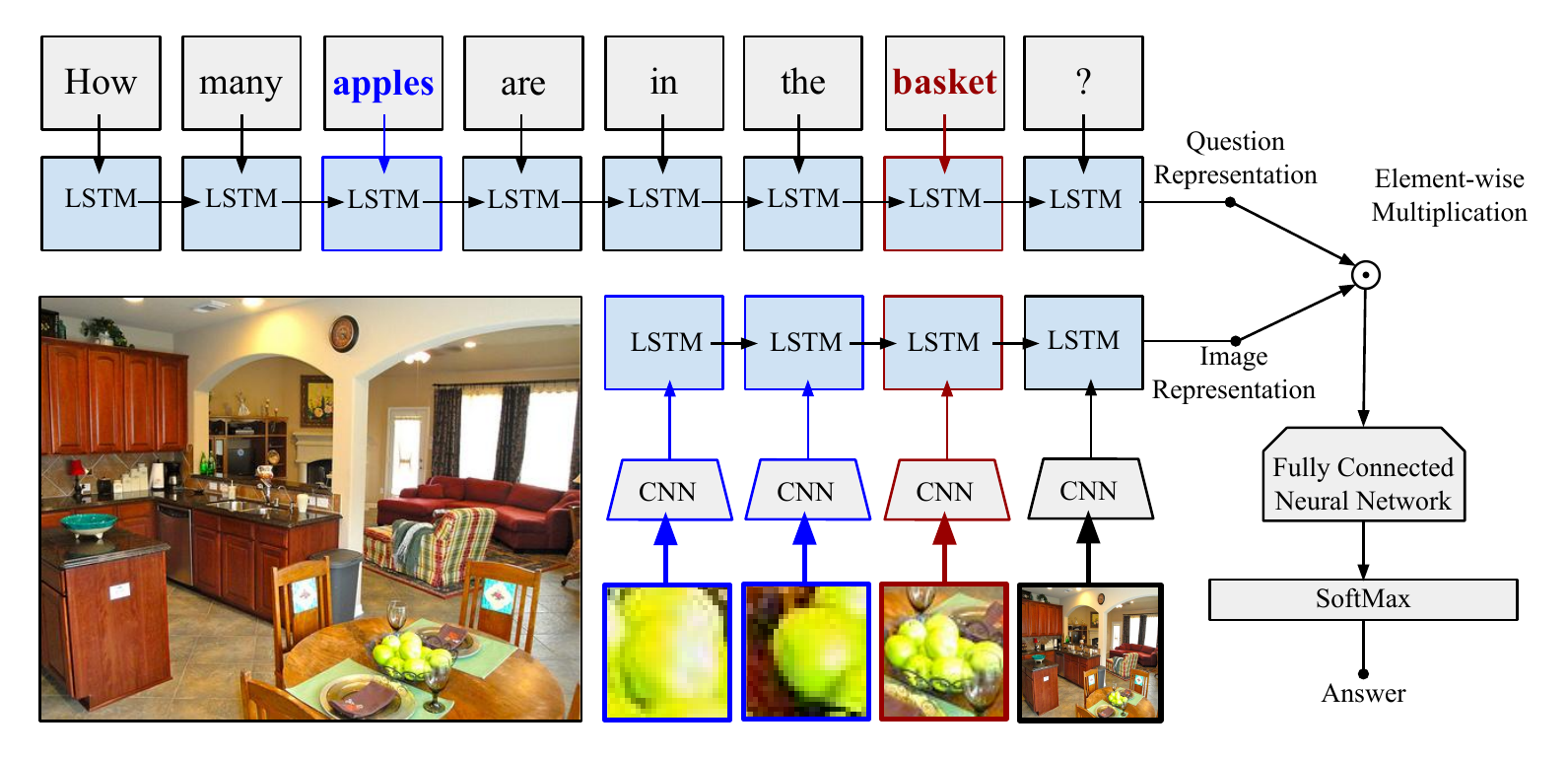}
\caption{Focused dynamic attention model diagram.}
\label{fig:move-attention}
\end{figure}

\section{Focused Dynamic Attention for VQA}\label{sec:model}
The FDA model is composed of question and image understanding components, attention mechanism, and a multimodal representation fusion network (Figure~\ref{fig:move-attention}).
In this section we describe them individually. 

\subsection{Question Understanding}
Following a common practice, our FDA model uses an LSTM network to encode the question in a vector representation \cite{hochreiter-97,malinowski-15,chen-15,noh-15}. 
The LSTM network learns to keep in its state the feature vectors of the important question words, and thus provides the question understanding component with a word attention mechanism.

\subsection{Image Understanding}
Following prior work \cite{ren-15,gao-15,malinowski-15}, we use a pre-trained convolutional neural network (CNN) to extract image feature vectors.
Specifically, we use the Deep Residual Networks model used in ILSVRC and COCO 2015 competitions, which won the 1\tsup{st} places in: ImageNet classification, ImageNet detection, ImageNet localization, COCO detection, and COCO segmentation \cite{he-15}.
We extract the weights of the layer immediately before the final SoftMax layer and regard them as visual features. 
We extract such features for the whole image (global visual features) and for the specific image regions (local visual features). 
However, contrary to the existing approaches, we employ an LSTM network to combine the local and global visual features into a joint representation. 

\subsection{Focused Dynamic Attention Mechanism}\label{sec:fda_model}
We introduce a focused dynamic attention mechanism that learns to focus on image regions related to the question words. 

The attention mechanism works as follows.
For each image object\footnote{During training we use the ground truth object bounding boxes and labels. At test time we use the precomputed bounding boxes from \cite{pont-15} and classify them with \cite{he-15} to obtain the object labels.} it uses word2vec word embeddings \cite{mikolov-13} to measure the similarity between the question words and the object label.
Next, it selects objects with similarity score greater than 0.5 and extracts the feature vectors of the objects bounding boxes with a pre-trained ResNet model \cite{he-15}.
Following the question word order, it feeds the LSTM network with the corresponding object feature vectors.
Finally, it feeds the LSTM network with the feature vector of the whole image and it uses the resulting LSTM state as a visual representation.
Thus, the attention mechanism enables the model to combine the local and global visual features into a single representation, necessary for answering complex visual questions.

Figure~\ref{fig:move-attention} illustrates the focused dynamic attention mechanism with an example.

\subsection{Multimodal Representation Fusion} 

We regard the final state of the two LSTM networks as a question and image representation.
We start fusing them into single representation by applying \emph{Tanh} on the question representation and \emph{ReLU}\footnote{Defined as $f(x) = \max(0, x)$.} on the image representation \footnote{Applying different activation functions gave slightly worse overall results}.  
We proceed by doing an element-wise multiplication of the two vector representations and the resulting vector is fed to a fully-connected neural network.
Finally a SoftMax layer classify the multimodal representation into one of the possible\footnote{We follow \cite{antol-15} and use the 1000 most common answers} answers.

\section{Evaluation}\label{sec:experiments} 
In this section we detail the model implementation and compare our model against the current state-of-the-art methods.

\begin{figure}[!ht] 
\includegraphics[width=\linewidth]{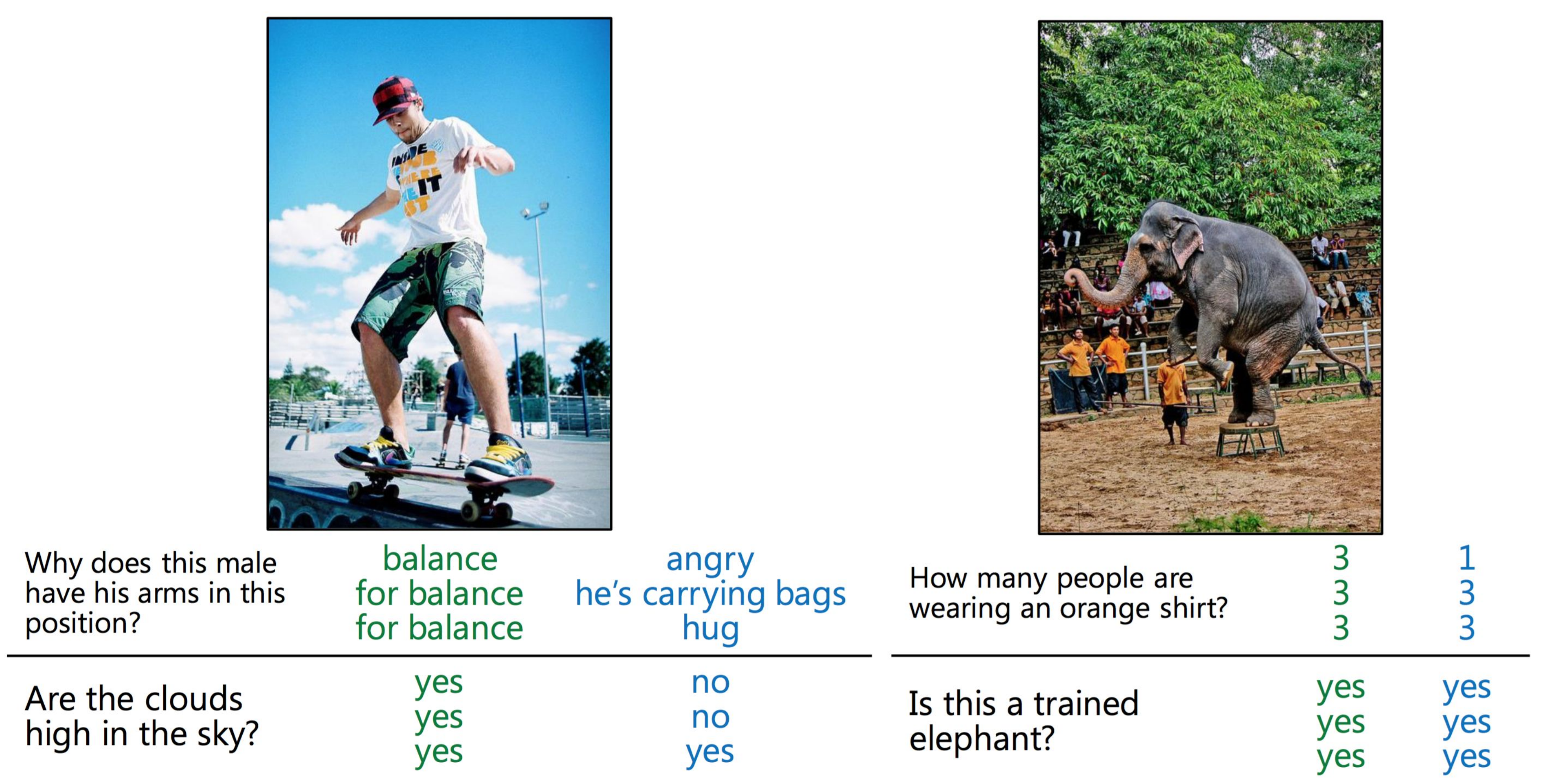}
\caption{Representative examples of questions (black), (a subset of the) answers given when looking at the image (green), and answers given when not looking at the image (blue) for two images from the VQA dataset. Examples provided by \cite{antol-15}.}
\label{fig:vqa-examples}
\end{figure}

\subsection{Dataset}
For all experiments we use the Visual Question Answering (VQA) dataset \cite{antol-15}, which is the largest and most complex image dataset for the visual question answering task.
The dataset contains three human posed questions and ten answers given by different human subjects, for each one of the 204,721 images found in the Microsoft COCO dataset \cite{lin-14}.
Figure~\ref{fig:vqa-examples} shows two representative examples found in the dataset.
The evaluation is done on following two test splits test-dev and test-std and on following two tasks:
\begin{itemize}
	\item An open-ended task, where the method should generate a natural language answer;
	\item A multiple-choice task, where for each question the method should chose one of the 18 different answers.
\end{itemize}

We evaluate the performance of all the methods in the experiments using the public evaluation server for fair evaluation.

\subsection{Baseline Model}
We compare our model against the baseline models provided by the VQA dataset authors \cite{lu-15}, which currently achieve the best performance on the test-standard split for the multiple-choice task.
The model, first described in \cite{antol-15}, is a standard implementation of an LSTM$+$CNN VQA model.
It uses an LSTM to encode the question and CNN features to encode the image.
To answer a question it multiplies the last LSTM state with the image CNN features and feeds the result into a SoftMax layer for classification into one of the 1,000 most common answers.
The implementation in \cite{lu-15} uses a deeper two layer LSTM network for encoding the question, and normalized image CNN features, which showed crucial for achieving the state-of-the-art.

\begin{table}[t!]

   \centering
    \caption{Comparison between the baselines from~\cite{antol-15}, the state-of-the-art models and our FDA model on VQA test-dev and test-standard data for the \textbf{open-ended} task. Results from most recent methods including CM \cite{jiang-15}, ACK \cite{wu-15}, iBOWIMG \cite{zhou-15}, DPPnet  \cite{noh-15}, D-NMN  \cite{andreas-16}, D-LSTM  \cite{lu-15},  and SAN \cite{yang-15} are provided and compared with. }
   \setlength{\tabcolsep}{.5em}
   \begin{tabular}{lllrrr}
      \hline
      \multicolumn{1}{c}{}&
      \multicolumn{4}{c}{test-dev}&
      \multicolumn{1}{r}{test-std}\\
      \cline{2-5}
      Method&All&Y/N&Other&Num&All\\
      \hline
      VQA& & & & &\\
      Question &48.09&75.66 &27.14 &36.70&-\\
      Image&28.13&64.01 &3.77 &0.42 &-\\
      Q+I&52.64&75.55 &37.37 &33.67 &-\\
      LSTM Q+I&53.74&78.94 &36.42 &35.24 &54.06\\
      \hline
      CM&52.62&78.33&35.93&34.46\\
      ACK&55.72&79.23 &40.08 &36.13 &55.98\\
      iBOWIMG&55.72&76.55 &42.62 &35.03 &55.89\\
      DPPnet&57.22&80.71 &41.71 &37.24 &57.36\\
      D-NMN&57.90&80.50 &43.10 &37.40 &58.00\\
      D-LSTM&-&-&-&-&58.16\\
      SAN&58.70&79.30 &46.10 &36.6 &58.90\\
      \hline
      FDA&\textbf{59.24}&81.14&45.77&36.16&\textbf{59.54}\\
   \end{tabular}
   \label{tbl:open-ended}
\end{table}

\begin{table}[t!]

   \centering
 \caption{Comparison between the baselines from~\cite{antol-15}, the state-of-the-art models and our FDA model on VQA test-dev and test-standard data for the \textbf{multiple-choice} task. Results from most recent methods including  WR  \cite{shih-15}, iBOWIMG \cite{zhou-15}, DPPnet  \cite{noh-15}, and D-LSTM  \cite{lu-15} are also shown for comparison.  }
   \setlength{\tabcolsep}{.5em}
   \begin{tabular}{lllrrr}
      \hline
      \multicolumn{1}{c}{}&
      \multicolumn{4}{c}{test-dev}&
      \multicolumn{1}{r}{test-std}\\
      \cline{2-5}
      Method&All&Y/N&Other&Num&All\\
      \hline
      VQA& & & & &\\
      Question &53.68&75.71 &38.64 &37.05&-\\
      Image&30.53&69.87 &3.76 &0.45 &-\\
      Q+I&58.97&75.59 &50.33 &34.35 &-\\
      LSTM Q+I&57.17&78.95 &43.41 &35.80 &57.57\\
      \hline
      WR&60.96& - & - & - &-\\
      iBOWIMG&61.68&76.68 &54.44 &38.94 &61.97\\
      DPPnet&62.48&80.79 &52.16 &38.94 &62.69\\
      D-LSTM&-&-&-&-&63.09\\
      \hline
      FDA&\textbf{64.01}&81.50&54.72&39.00&\textbf{64.18}\\
   \end{tabular}
   \label{tbl:multi}
\end{table}

\subsection{Model Implementation and Training Details}
We transform the question words into a vector form by multiplying one-hot vector representation with a word embedding matrix.
The vocabulary size is 12,602 and the word embeddings are 300 dimensional.
We feed a pre-trained ResNet network \cite{he-15} and use the 2,048 dimensional weight vector of the layer before the last fully-connected layer.

The word and image vectors are feed into two separate LSTM networks.
The LSTM networks are standard implementation of one layer LSTM network \cite{hochreiter-97}, with a 512 dimensional state vector.
The final state of the question LSTM is passed through \emph{Tanh}, while the final state of the image LSTM is passed through {ReLU}\footnote{Defined as $f(x) = \max(0, x)$.}.
We do element-wise multiplication on the resulting vectors, to obtain a multimodal representation vector, which is then fed to a fully-connected neural network.

\subsection{Model Evaluation and Comparison}

We compare our model with the baselines provided by the VQA authors~\cite{antol-15}.
The results for the open-ended task are listed in Table~\ref{tbl:open-ended} and the results for the multiple-choice task are given in Table~\ref{tbl:multi}.
In the tables, the ``Question'' and ``Image'' baselines are only using the question words and the image, respectively.
The ``Q+I'' is a baseline that combines the two, but do not use an LSTM network.  
``LSTM Q+I'' and ``D-LSTM'' are LSTM models, with one and two layers accordingly. 
Comparing the performance of baselines we can observe the accuracy increase with the addition of information from each modality. 

From Table~\ref{tbl:open-ended}, one can observe that our proposed FDA model achieves the best performance on this benchmark dataset.
It outperforms the state-of-the-art (SAN) with a margin of around $0.6\%$.
The SAN model also employs attention to focus on specific regions.
However, their attention model (without access to the automatically generated bounding boxes) is focusing on more spread regions which may include cluttered and noisy background.
In contrast, FDA only focuses on the selected regions and extracts more clean information for answering the questions.
This is the main reason that FDA can outperform SAN although these two methods are both based on attention models.

The advantage of employing focused dynamic attention in FDA is more significant when solving the multiple-choice VQA problems. 
From Table \ref{tbl:multi}, one can observe that our proposed FDA model achieves the best ever performance on the VQA dataset.
In particular, it improves the performance of the state-of-the-art (D-LSTM) by a margin of $1.1\%$ which is quite significant for this challenging task.
The D-LSTM method employs a deeper network to enhance the discriminative capacity of the visual features.
However, they do not identify the informative regions for answering the questions.
In contrast, FDA incorporates the automatic region localization by employing a question-driven attention model.
This is helpful for filtering out irrelevant noise, and establishing the correspondence between regions and candidate answers.
Thus FDA gives substantial performance improvement.

\subsection{Qualitative Results}

We qualitatively evaluate our model on a set of examples where complex reasoning and focusing on the relevant local visual features are needed for answering the question correctly.  

\begin{figure*}
\begin{center}
\begin{tabular}{c@{}c}
   \includegraphics[clip=true,width=0.48\textwidth,keepaspectratio]{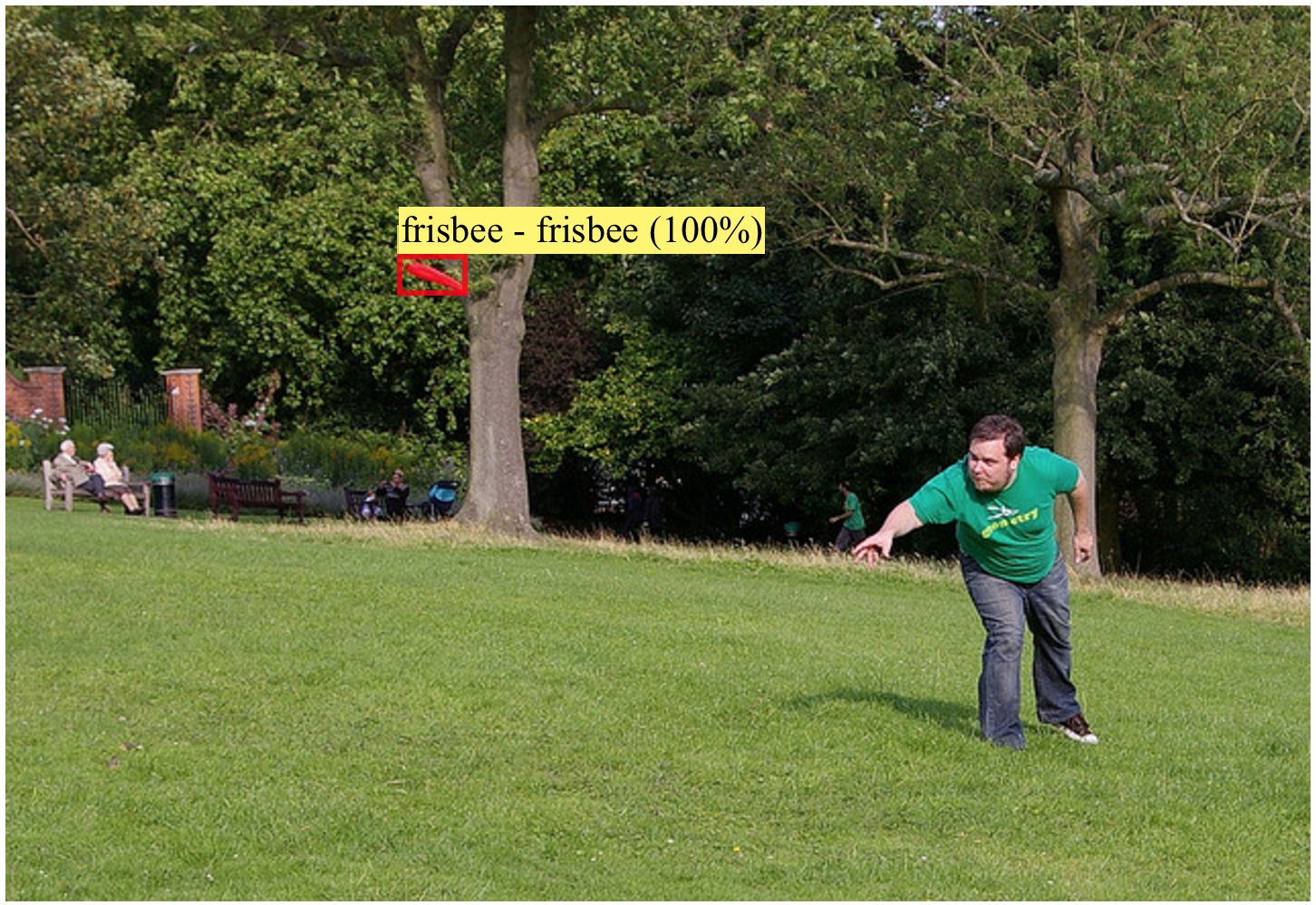} &
   \includegraphics[clip=true,width=0.48\textwidth,keepaspectratio]{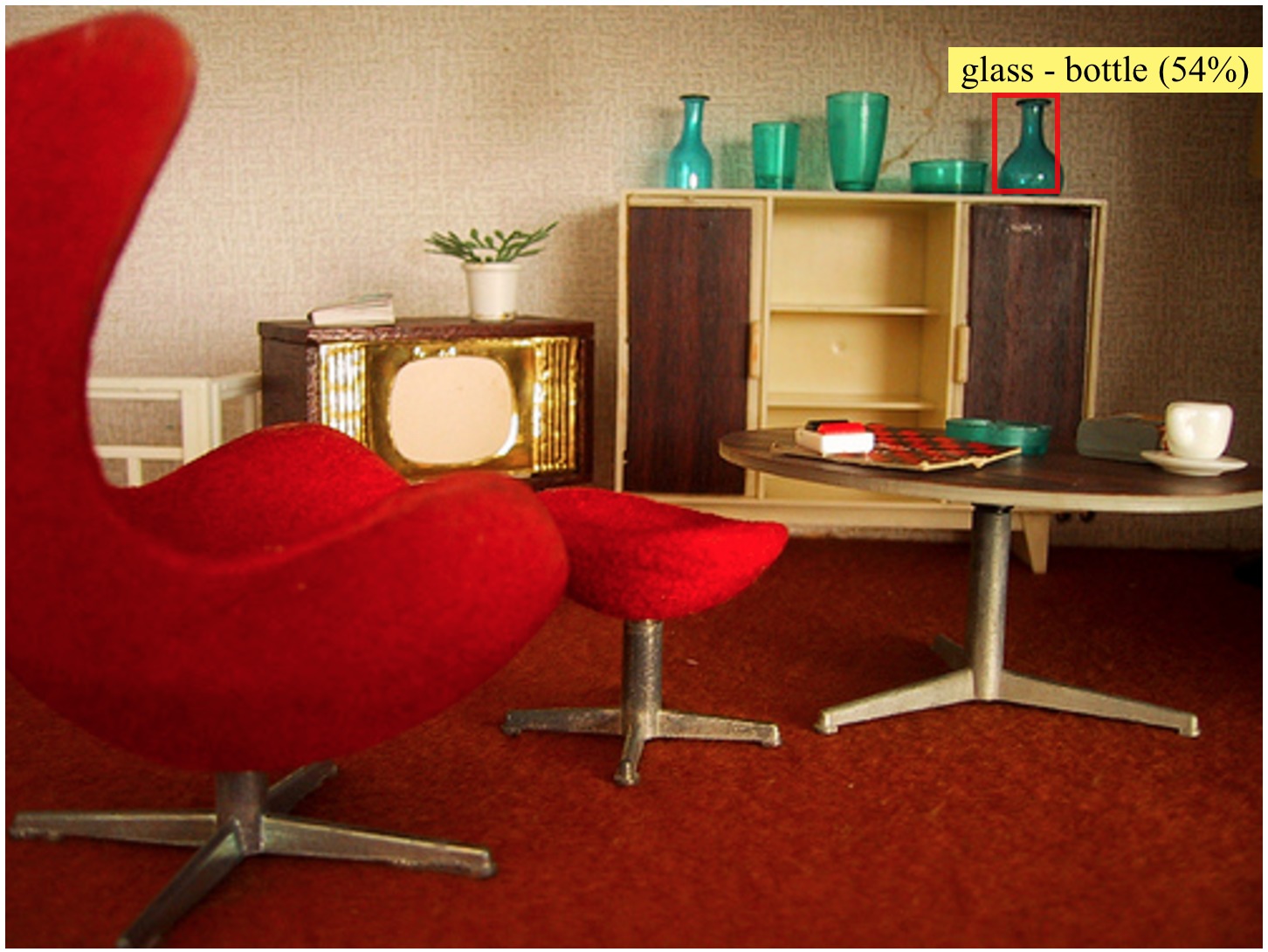} \\
\multicolumn{1}{m{5.8cm}}{\small{
      What color is the \textbf{frisbee}?\newline - Red.\newline
}} &
\multicolumn{1}{m{5.8cm}}{\small{
      What color are the \textbf{glass} items?\newline - Green.\newline
}} \\
   \includegraphics[clip=true,width=0.48\textwidth,keepaspectratio]{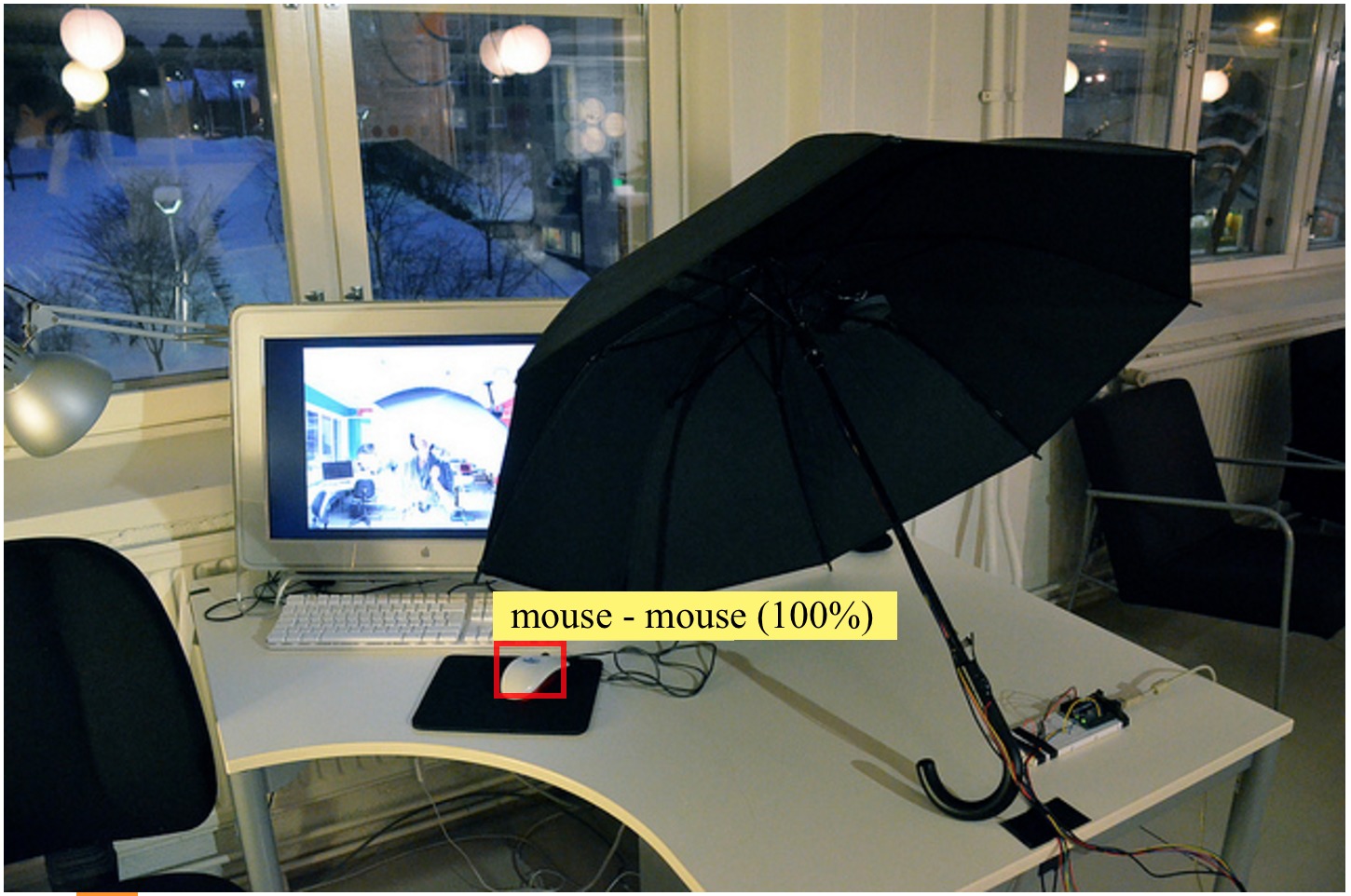} &
   \includegraphics[clip=true,width=0.48\textwidth,keepaspectratio]{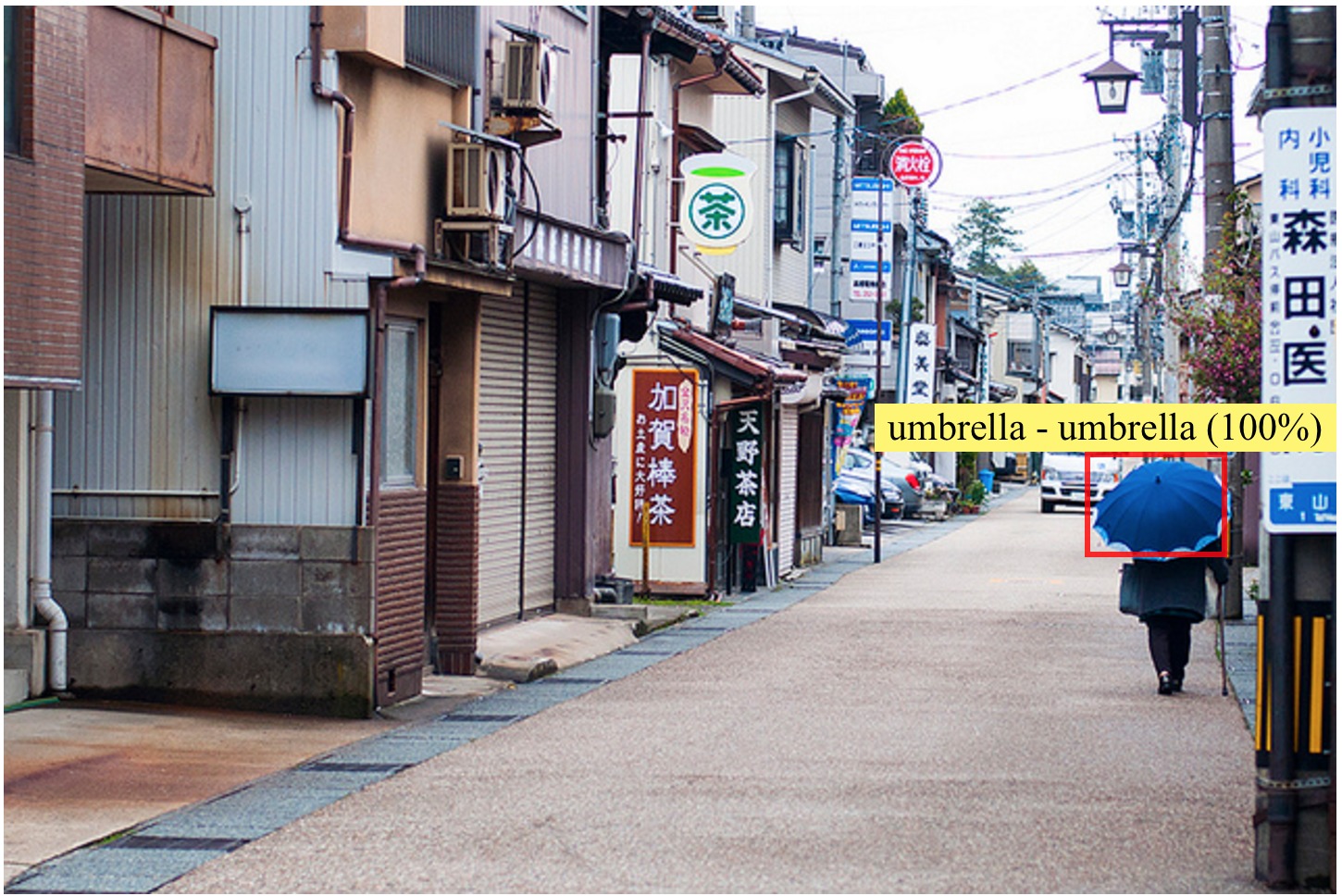} \\
\multicolumn{1}{m{5.8cm}}{\small{
      What color is the \textbf{mouse}?\newline- White.
}} &
\multicolumn{1}{m{5.8cm}}{\small{
     What color is the lady's \textbf{umbrella}?\newline - Blue
}} \\
\end{tabular}
\end{center}
\caption{Representative examples where focusing on the question related objects helps FDA answer ``What color'' type of questions.
The question words in bold have been matched with an image region.
The yellow region caption box contains the question word, followed by the region label, and in parenthesis their cosine similarity (see Section~\ref{sec:fda_model} for more details).
}
\label{fig:color-examples}
\end{figure*}

Figure~\ref{fig:color-examples} shows particularly difficult examples (the predominant image color is \textbf{not} the correct answer) of ``What color'' type of questions.
But, by focusing on the question related image regions, the FDA model is still able to produce the correct answer.

\begin{figure*}
\begin{center}
\begin{tabular}{c@{}c}

   \includegraphics[clip=true,width=0.48\textwidth,height=0.35\textheight,keepaspectratio]{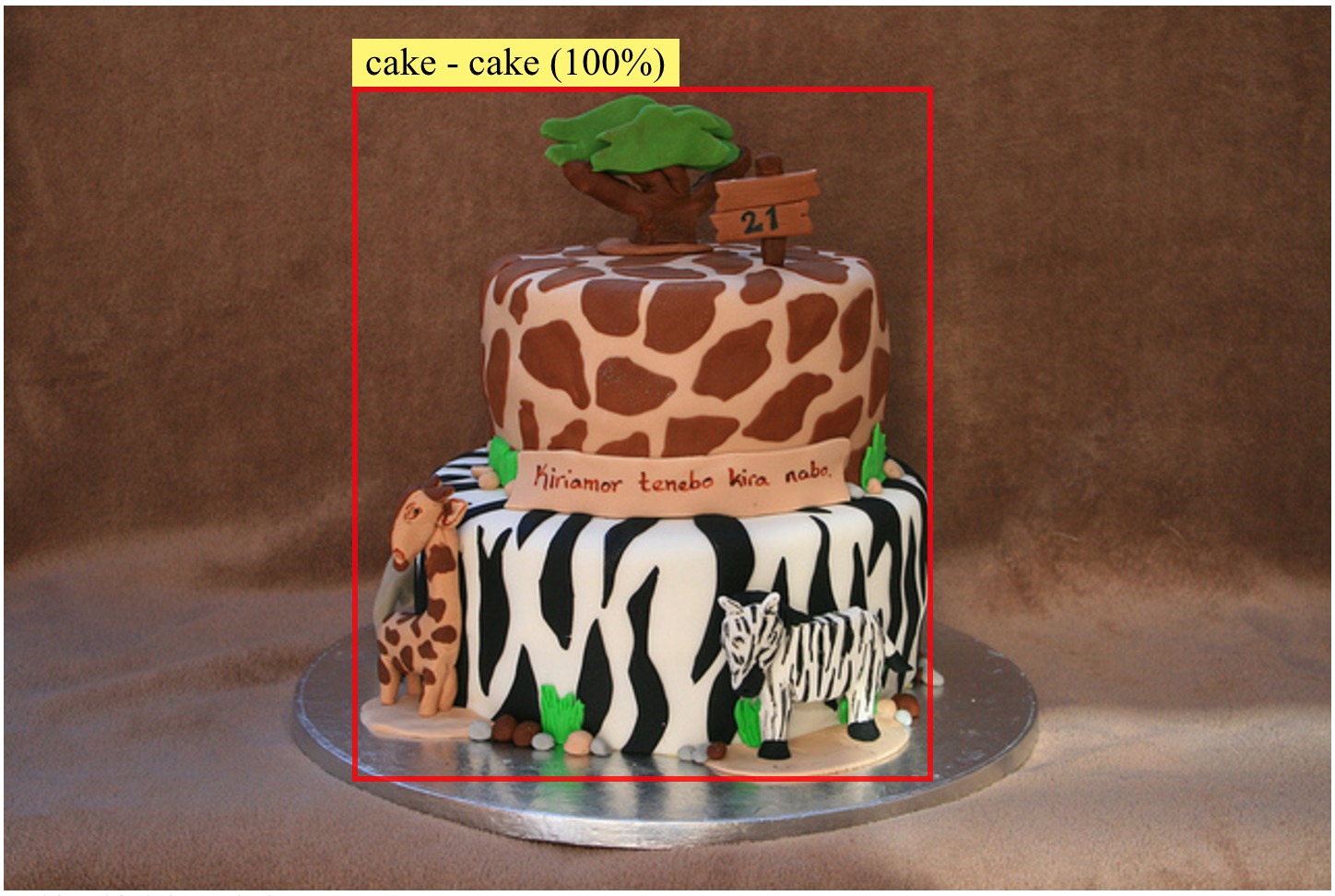} &
   \includegraphics[clip=true,width=0.48\textwidth,height=0.35\textheight,keepaspectratio]{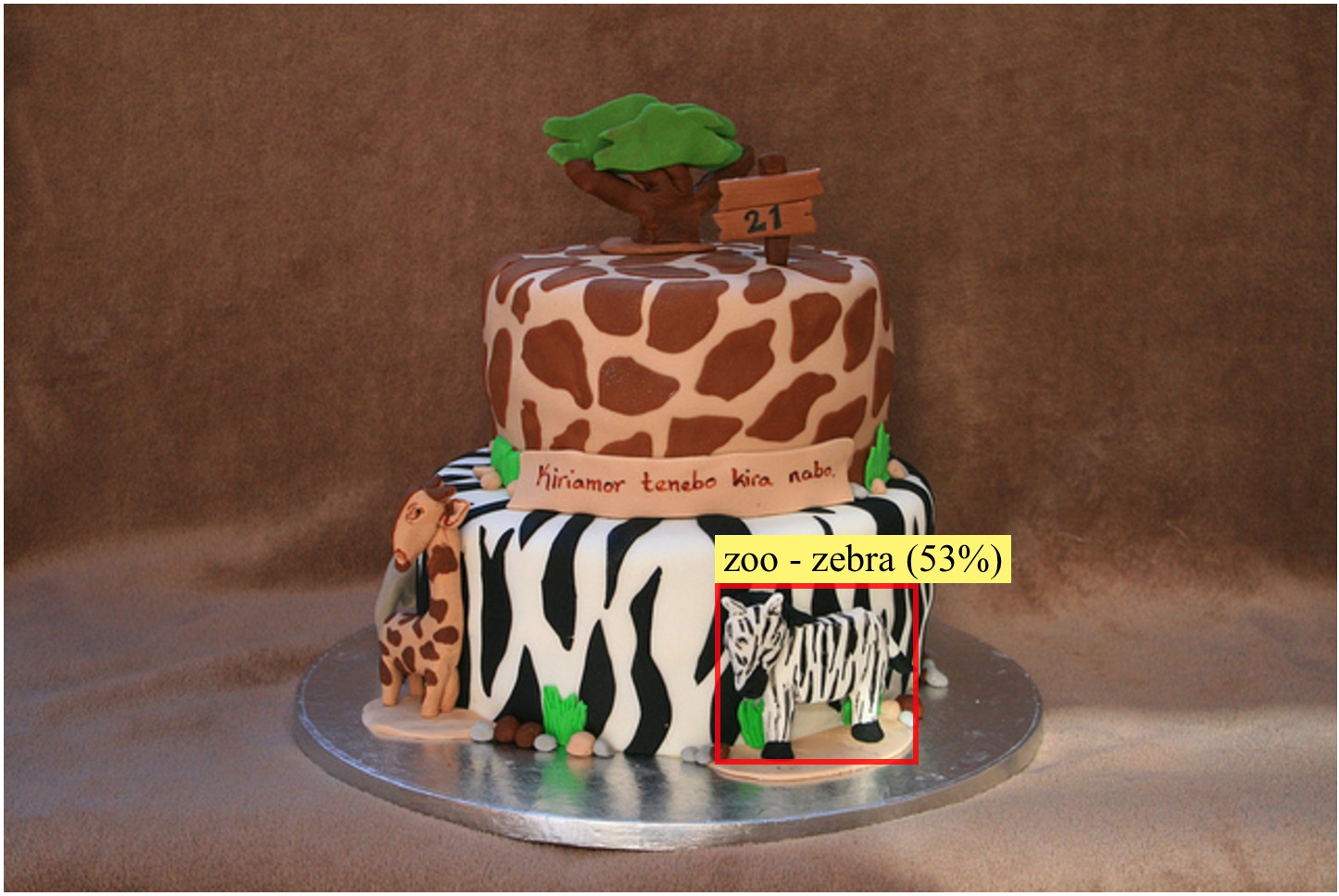} \\
\multicolumn{1}{m{5.8cm}}{\small{
      Is this a birthday \textbf{cake}? \newline- Yes.
}} &
\multicolumn{1}{m{5.8cm}}{\small{
      Is someone in all likelihood, a \textbf{zoo} fancier?  - Yes.
}} \\
   \includegraphics[clip=true,width=0.48\textwidth,height=0.35\textheight,keepaspectratio]{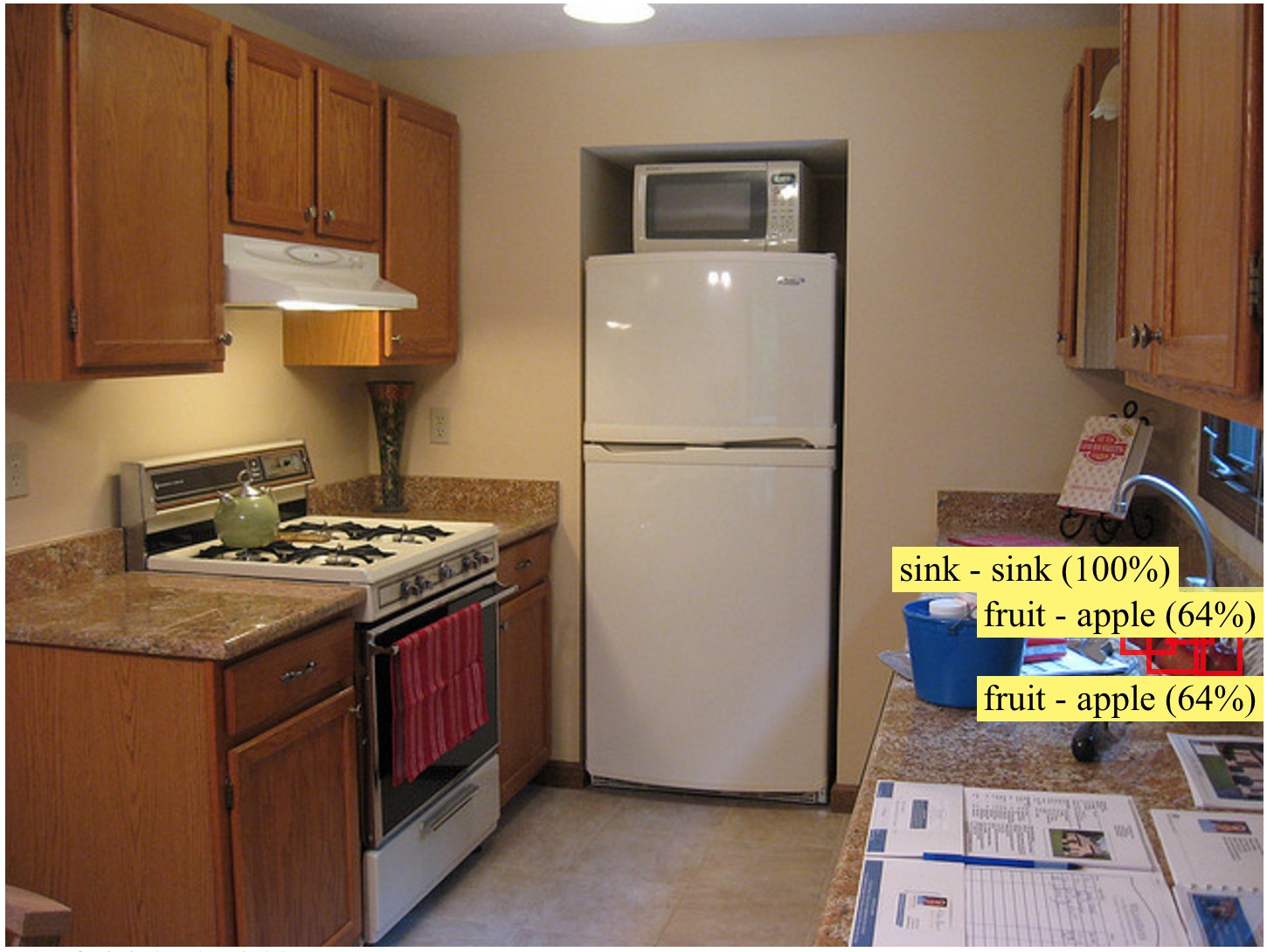} &
   \includegraphics[clip=true,width=0.48\textwidth,height=0.35\textheight,keepaspectratio]{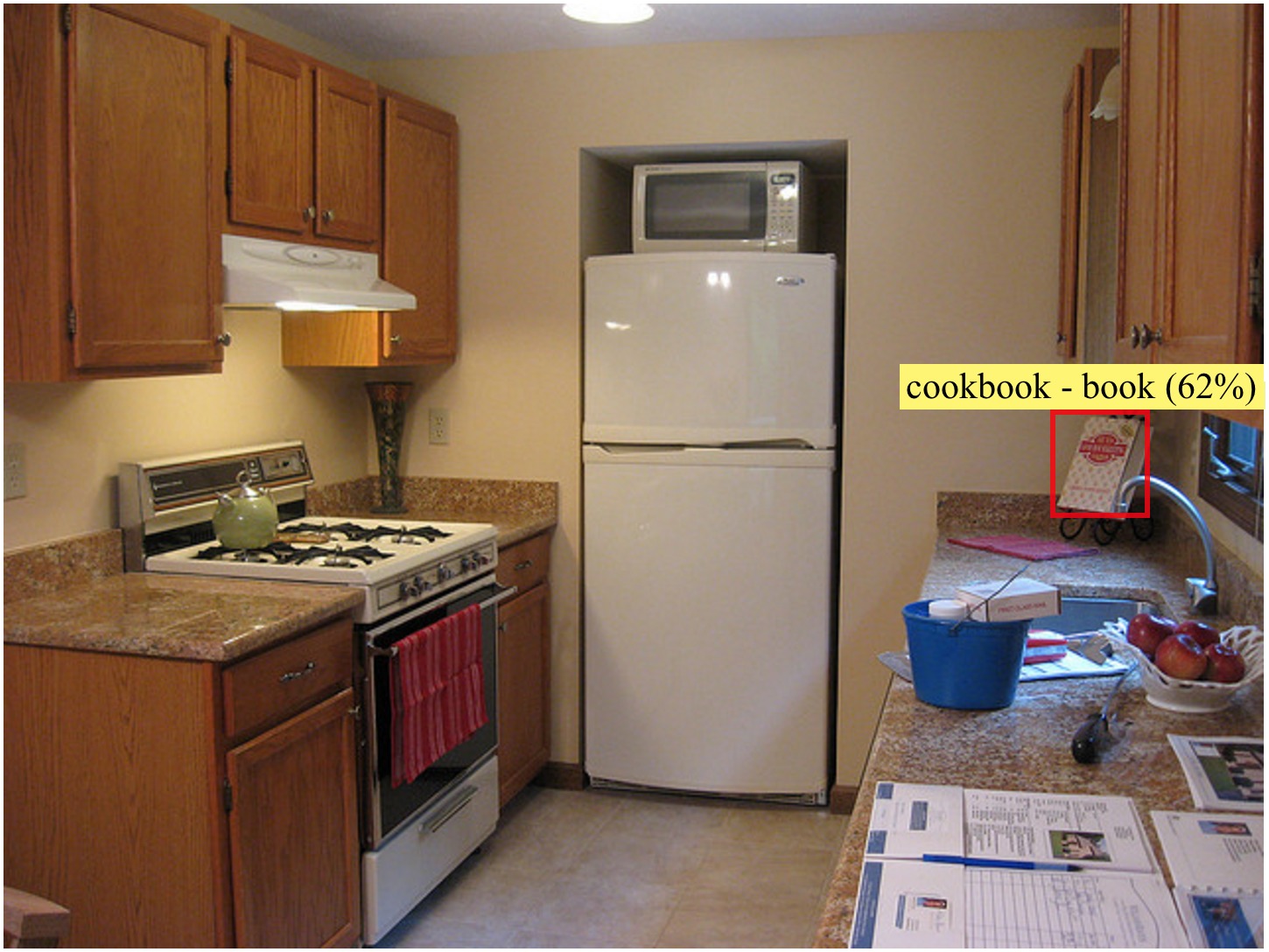} \\
\multicolumn{1}{m{5.8cm}}{\small{
      What \textbf{fruit} is by the \textbf{sink}? \newline- Apples.
}} &
\multicolumn{1}{m{5.8cm}}{\small{
      Is there a \textbf{cookbook} in the picture?\newline - Yes.     
}} \\
   \includegraphics[clip=true,width=0.48\textwidth,height=0.35\textheight,keepaspectratio]{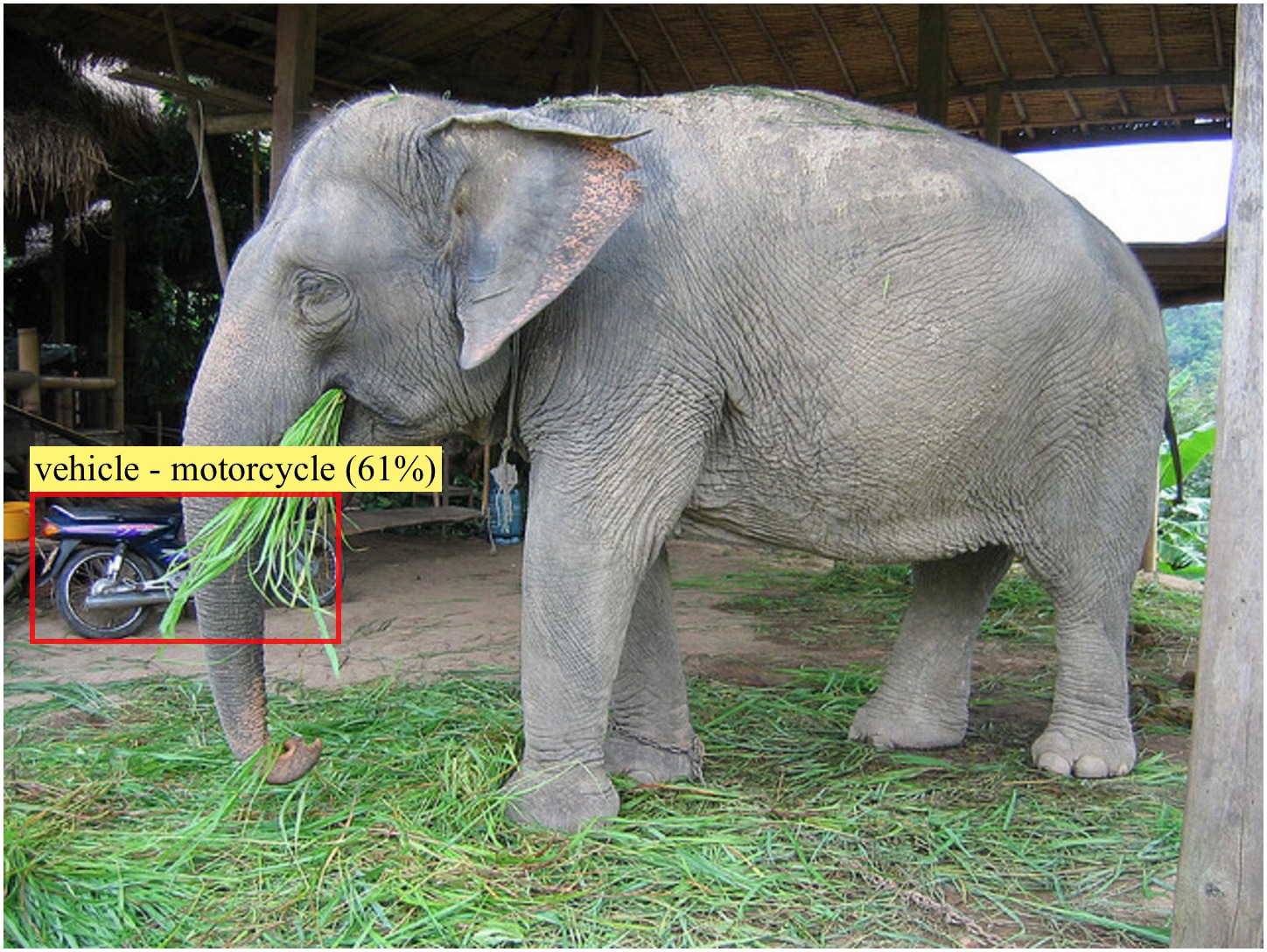} &
   \includegraphics[clip=true,width=0.48\textwidth,height=0.35\textheight,keepaspectratio]{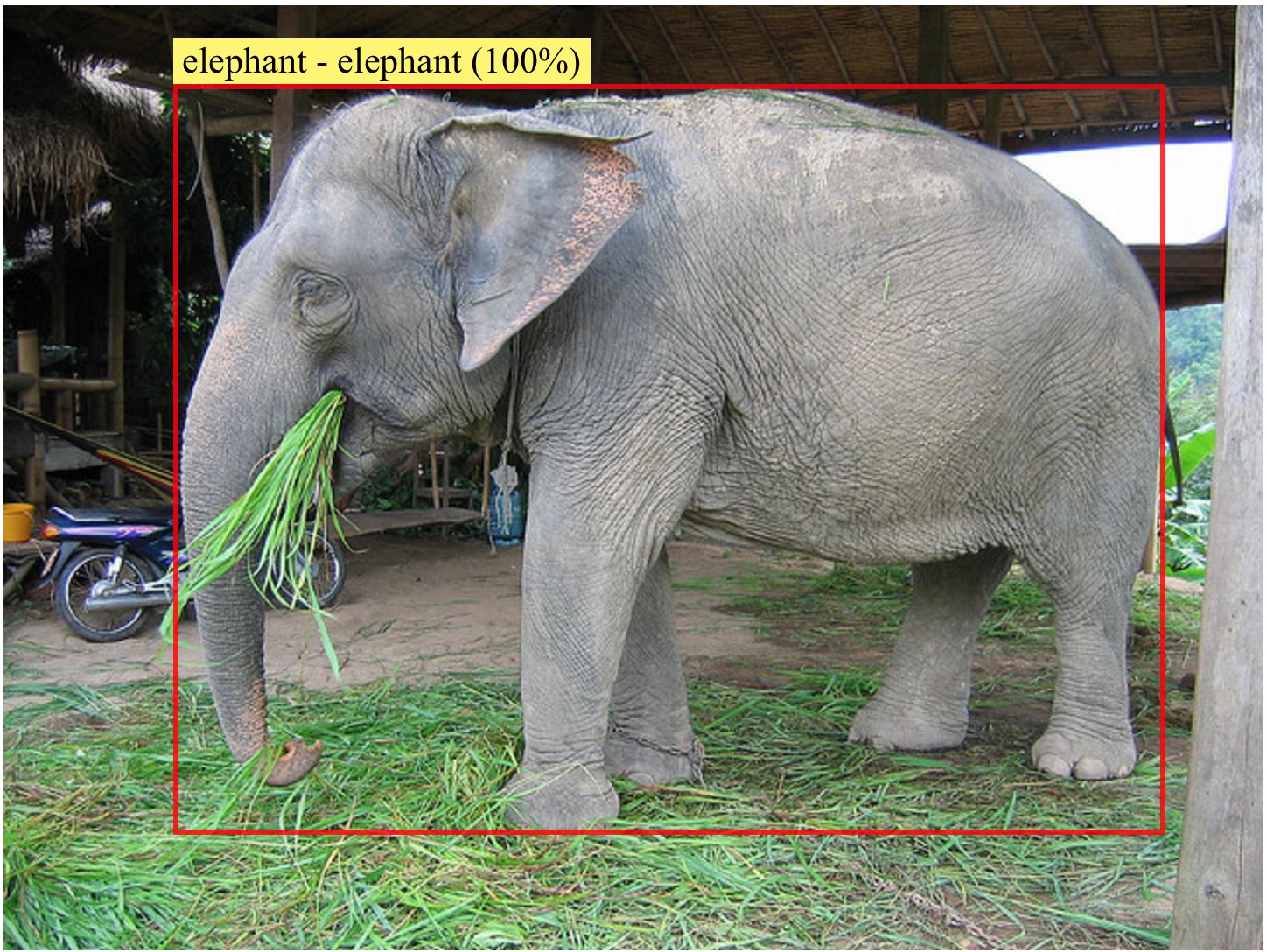} \\
\multicolumn{1}{m{5.8cm}}{\small{
      What type of \textbf{vehicle} is pictured?\newline -  Motorcycle.
}} &
      \multicolumn{1}{m{5.8cm}}{\small{
            Does the \textbf{elephant} have tusks? \newline- No.
}} \\
\end{tabular}
\end{center}
\caption{Representative examples where the model focuses on different regions from the \textbf{same} image, depending on the question.
The question words in bold have been matched with an image region.
The yellow region caption box contains the question word, followed by the region label, and in parenthesis their cosine similarity (see Section~\ref{sec:fda_model} for more details).
}
\label{fig:same-image-examples}
\end{figure*}

In Figure~\ref{fig:same-image-examples} we show examples where the model focuses on different regions from the \textbf{same} image, depending on the words in the question.
Focusing on the right image region is crucial when answering unusual questions for an image (Row~1), questions about small image objects (Row~2), or when the most dominant image object partly occludes the question related region and can lead to a wrong answer (Row~3).

\begin{figure*}
\begin{center}
\begin{tabular}{c@{}c}

   \includegraphics[clip=true,width=0.48\textwidth,height=0.3\textheight,keepaspectratio]{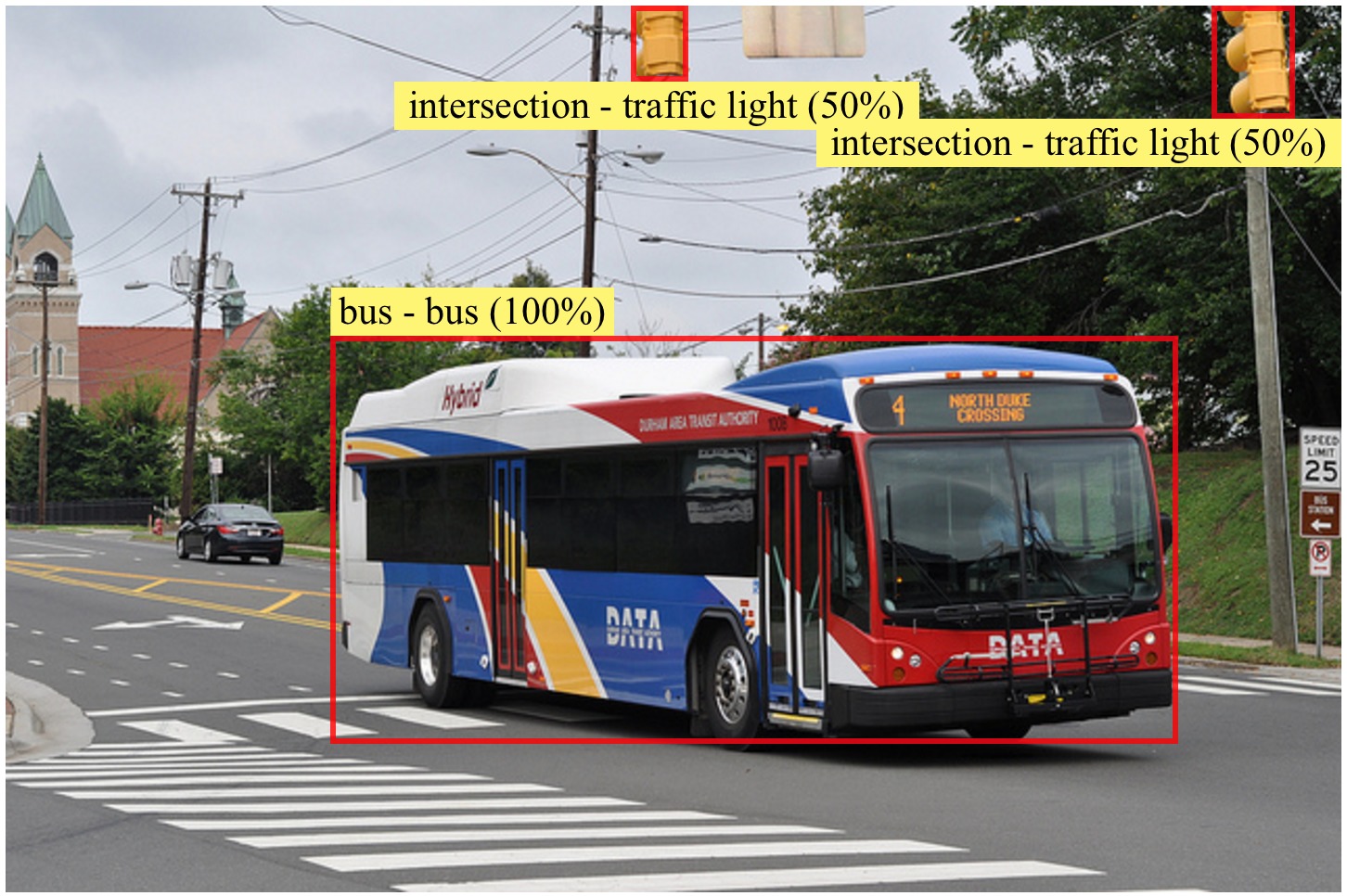} &
   \includegraphics[clip=true,width=0.48\textwidth,height=0.35\textheight,keepaspectratio]{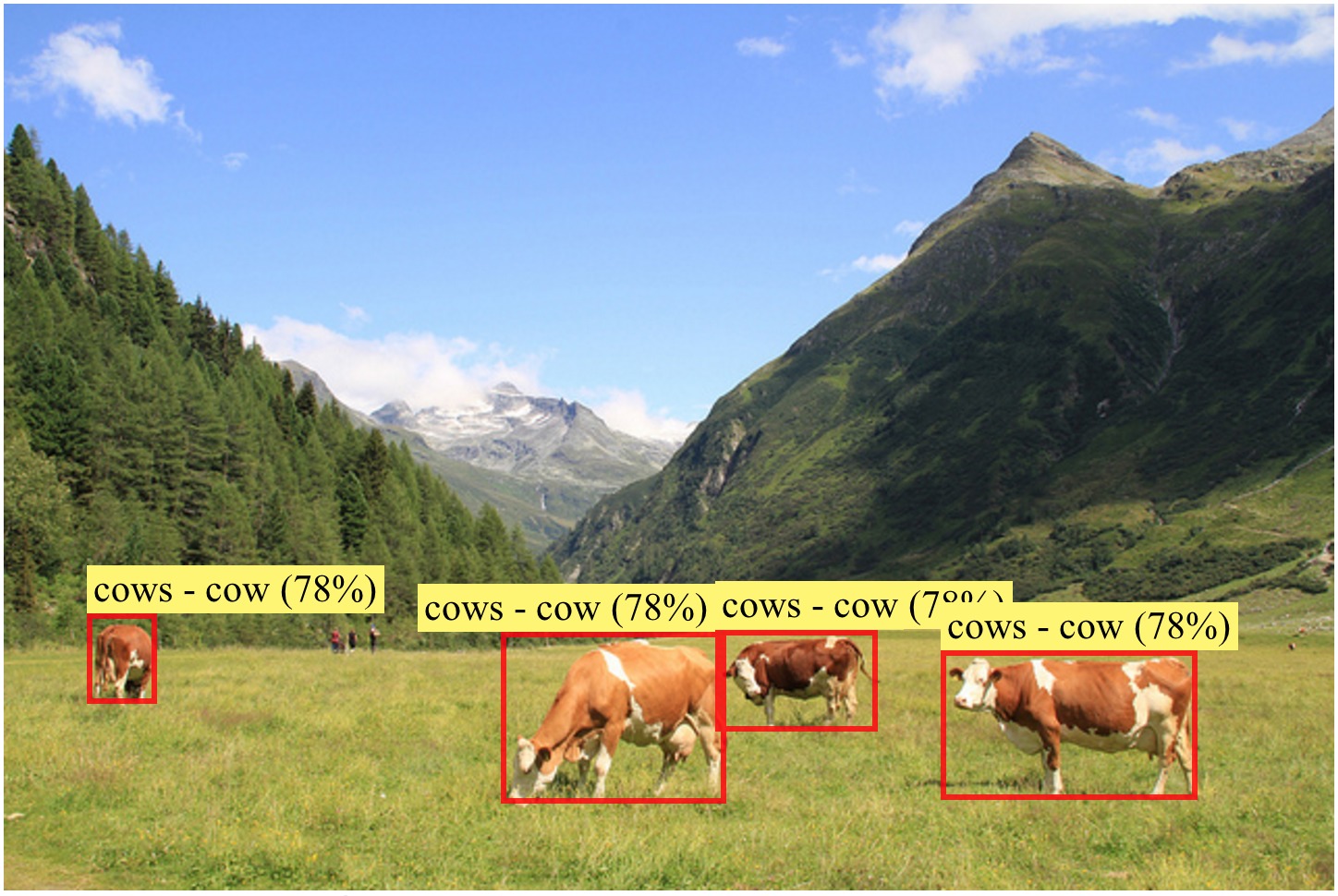} \\
\multicolumn{1}{m{5.8cm}}{\small{
      Is the \textbf{bus} in the middle of the \textbf{intersection}? - Yes.
}} &
\multicolumn{1}{m{5.8cm}}{\small{
      How many \textbf{cows} are present?\newline- 4.
}} \\
   \includegraphics[clip=true,width=0.48\textwidth,height=0.35\textheight,keepaspectratio]{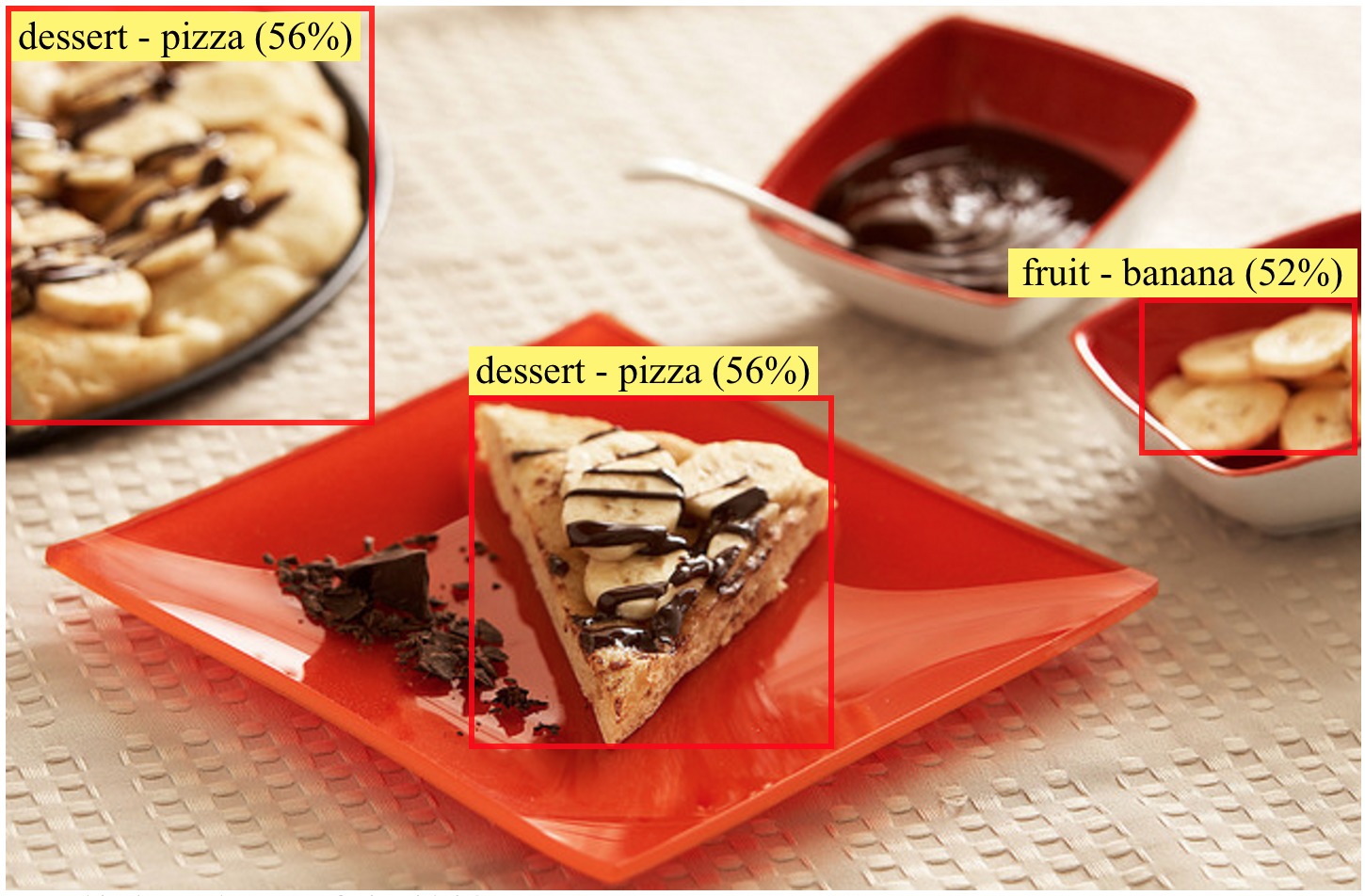} &
   \includegraphics[clip=true,width=0.48\textwidth,height=0.35\textheight,keepaspectratio]{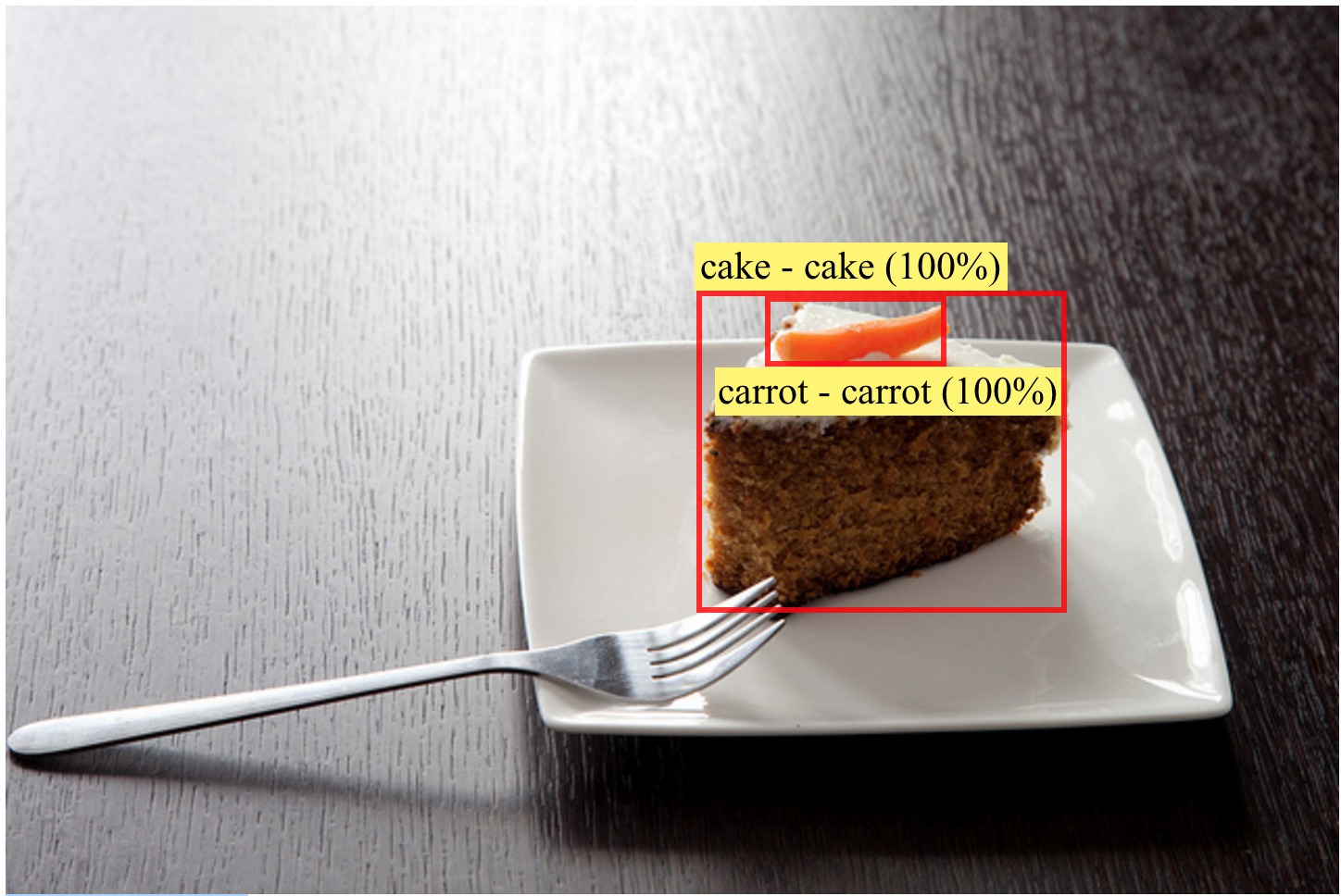} \\
\multicolumn{1}{m{5.5cm}}{\small{
      Does this \textbf{dessert} have any \textbf{fruit} with it? - Yes.
}} &
\multicolumn{1}{m{5.8cm}}{\small{
      Is this a \textbf{carrot} \textbf{cake}?\newline - Yes.
}} \\
   \includegraphics[clip=true,width=0.48\textwidth,height=0.35\textheight,keepaspectratio]{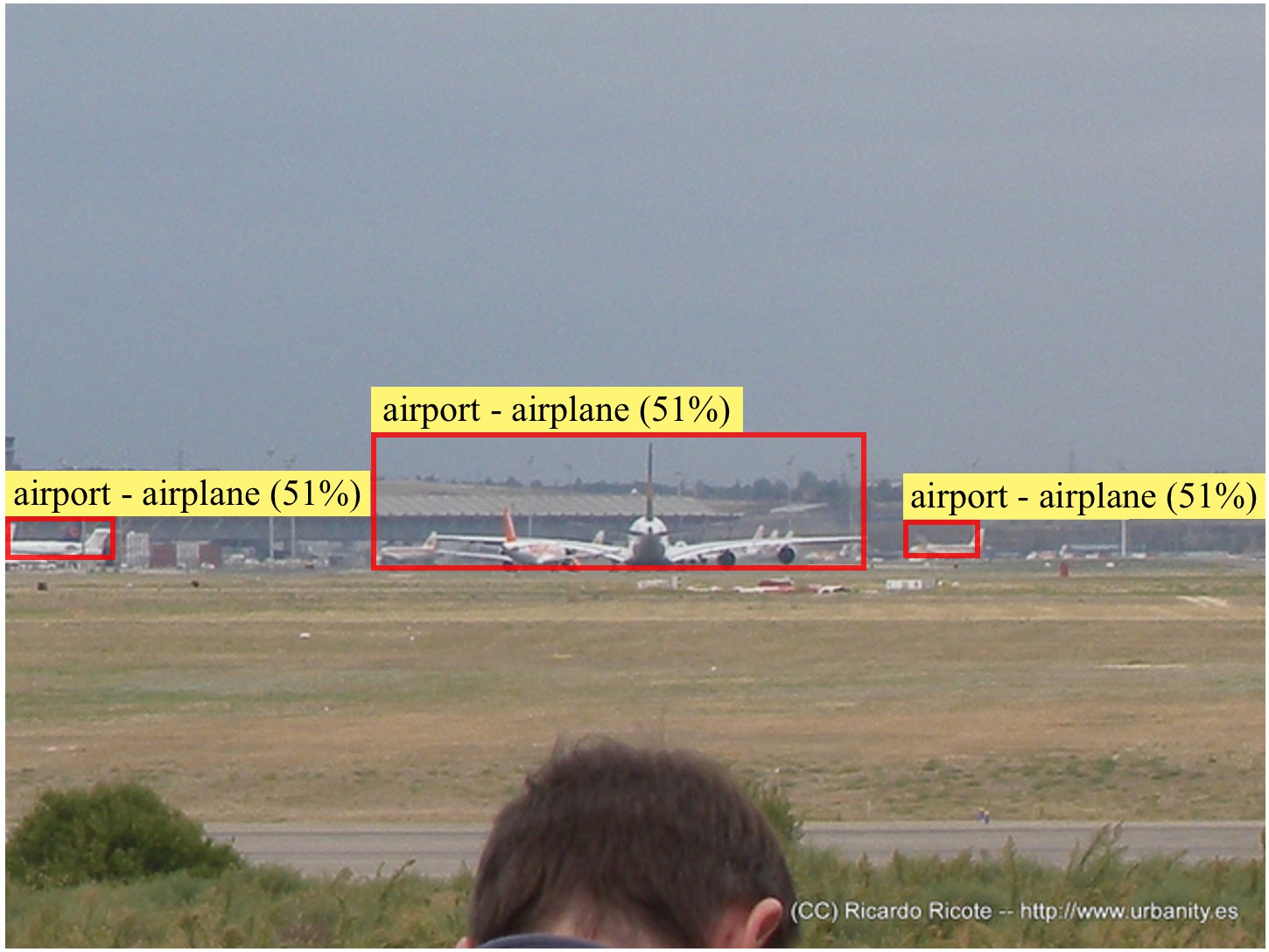} &
   \includegraphics[clip=true,width=0.48\textwidth,height=0.3\textheight,keepaspectratio]{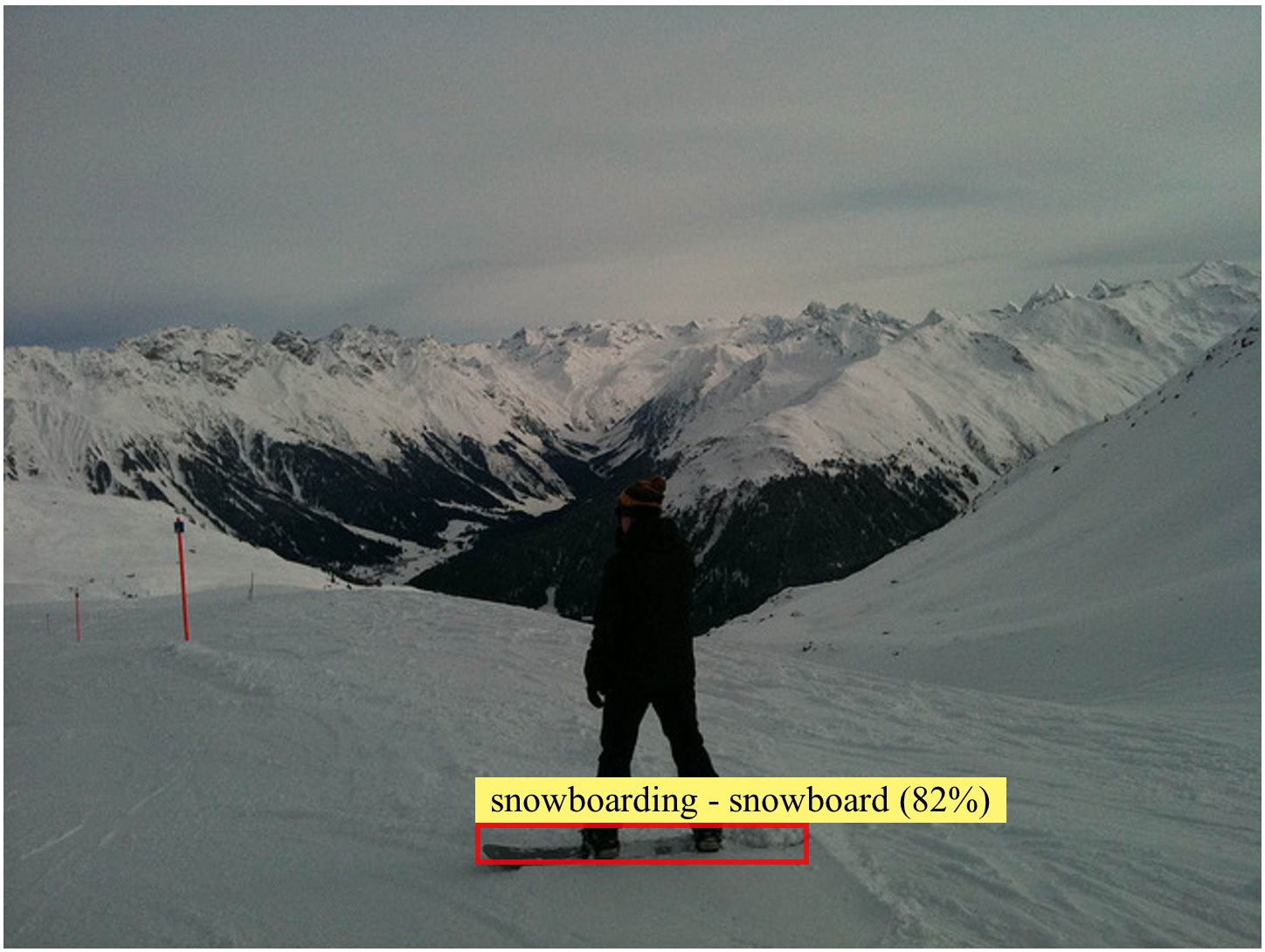} \\
\multicolumn{1}{m{5.8cm}}{\small{
      Is this an \textbf{airport}?\newline - Yes.
}} &
      \multicolumn{1}{m{5.8cm}}{\small{
            Is the individual skiing or \textbf{snowboarding}? - Snowboarding.
}} \\
\end{tabular}
\end{center}
\caption{Representative examples of questions that require image object identification.
The question words in bold have been matched with an image region.
The yellow region caption box contains the question word, followed by the region label, and in parenthesis their cosine similarity (see Section~\ref{sec:fda_model} for more details).
}
\label{fig:complex-examples}
\end{figure*}

Representative examples of questions that require image object identification are shown in Figure~\ref{fig:complex-examples}.
We can observe that the focused attention enables the model to answer complex questions (Row~1, left) and counting questions (Row~1, right). 
The question guided image object identification greatly simplifies the answering of questions like the ones shown in Row~2 and Row~3.


\section{Conclusion}\label{sec:conclusion}
In this work, we proposed a novel Focused Dynamic Attention (FDA) model to solve the challenging VQA problems.
FDA is built upon a generic object-centric attention model for extracting question related  visual features from an image as well as a stack of multiple LSTM layers for feature fusion.
By only focusing on the identified regions specific for proposed questions, FDA was shown to be able to filter out overwhelming irrelevant informations from cluttered background or other regions, and thus substantially improved the quality of visual  representations in the sense of answering proposed questions.
By fusing cleaned regional representation, global context and question representation via LSTM layers, FDA provided significant performance improvement over baselines on the VQA benchmark datasets, for both the open-ended and multiple-choices VQA tasks.
Excellent performance of FDA clearly demonstrates its stronger ability of modeling visual contents and also verifies paying more attention to visual part in VQA tasks could essentially improve the overall performance.
In the future, we are going to further explore along this research line and investigate different attention methods for visual information selection as well as better reasoning model  for interpreting the relation between visual contents and questions.
\bibliographystyle{splncs}
\bibliography{egbib}
\end{document}